\documentclass[letterpaper]{article} 
\usepackage{aaai2026}  
\usepackage{times}  
\usepackage{helvet}  
\usepackage{courier}  
\usepackage[hyphens]{url}  
\usepackage{graphicx} 
\usepackage{natbib}  
\usepackage{caption} 
\usepackage{algorithm}
\usepackage{algorithmic}

\usepackage{newfloat}
\usepackage{listings}
\DeclareCaptionStyle{ruled}{labelfont=normalfont,labelsep=colon,strut=off} 
\floatstyle{ruled}
\newfloat{listing}{tb}{lst}{}
\floatname{listing}{Listing}
\usepackage{multirow}
\usepackage{bm}
\usepackage{amsmath}  
\usepackage{amssymb}  
\usepackage{booktabs}

\title{BAT: Learning Event-based Optical Flow with Bidirectional Adaptive Temporal Correlation}
\author{
    Gangwei Xu\textsuperscript{\rm 1,2}, Haotong Lin\textsuperscript{\rm 3}, Zhaoxing Zhang\textsuperscript{\rm 1}, Hongcheng Luo\textsuperscript{\rm 2}, Haiyang Sun\textsuperscript{\rm 2}, Xin Yang\textsuperscript{\rm 1}\thanks{Corresponding author.}\\
}
\affiliations{
    \textsuperscript{\rm 1}Huazhong University of Science and Technology \quad \textsuperscript{\rm 2}Xiaomi EV \quad \textsuperscript{\rm 3}Zhejiang University\\

}

\begin{document}

\maketitle

\begin{abstract}
Event cameras deliver visual information characterized by a high dynamic range and high temporal resolution, offering significant advantages in estimating optical flow for complex lighting conditions and fast-moving objects. Current advanced optical flow methods for event cameras largely adopt established image-based frameworks. However, the spatial sparsity of event data limits their performance. In this paper, we present BAT, an innovative framework that estimates event-based optical flow using bidirectional adaptive temporal correlation. BAT includes three novel designs: 1) a bidirectional temporal correlation that transforms bidirectional temporally dense motion cues into spatially dense ones, enabling accurate and spatially dense optical flow estimation; 2) an adaptive temporal sampling strategy for maintaining temporal consistency in correlation; 3) spatially adaptive temporal motion aggregation to efficiently and adaptively aggregate consistent target motion features into adjacent motion features while suppressing inconsistent ones. Our BAT achieves state-of-the-art performance on the DSEC-Flow benchmark, outperforming existing methods by a large margin while also exhibiting sharp edges and high-quality details. Our BAT can accurately predict future optical flow using only past events, significantly outperforming E-RAFT's warm-start approach.
\end{abstract}

\begin{links}
    \link{Code}{https://github.com/gangweix/BAT}
\end{links}

\section{Introduction}

Optical flow has been an enduring fundamental task in the field of computer vision, which can be applied to a broad range of practical tasks, including video reconstruction~\cite{tulyakov2021time, tulyakov2022time,xu2024hdrflow} and visual odometry~\cite{zhang2024leveraging, shiba2024secrets}, and robotics~\cite{xu2025pixel,liang2025parameter,liang2025sood++,li2025recogdrive,lin2025prompting}. Image-based optical flow estimation~\cite{flownet,raft,sea-raft} has demonstrated remarkable performance on public benchmarks~\cite{sintel,kitti15,spring}. Nonetheless, these methods struggle with motion blur and over-exposed or under-exposed regions due to the limited dynamic range of conventional image sensors. Event cameras present a promising alternative for optical flow estimation. Compared to image-based cameras, event cameras continuously capture asynchronous brightness changes, providing high temporal resolution without motion blur (microseconds), along with a high dynamic range and low power consumption. Therefore, event-based optical flow estimation is more advantageous for fast-moving objects, high dynamic range scenes, and resource-constrained devices.

\begin{figure}[t]
\centering
{\includegraphics[width=1.0\linewidth]{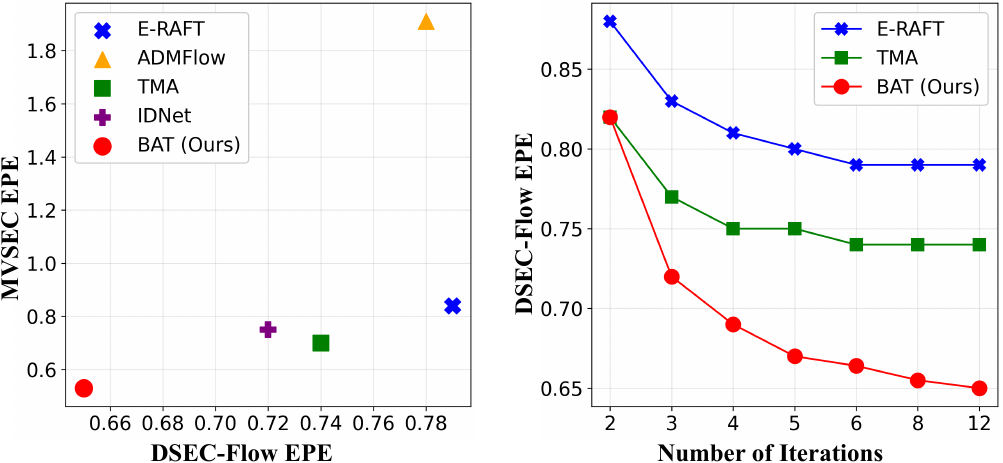}}
\caption{\textbf{Left:} Comparison with state-of-the-art event-based flow methods on DSEC-Flow and MVSEC benchmarks. Our BAT achieves the highest accuracy among all published methods. \textbf{Right:} Compared to E-RAFT and TMA, our BAT shows much higher efficiency in iterative optimization. Additionally, the aforementioned methods converge at a high error rate quickly due to limited unidirectional motion cues. In contrast, our BAT fully utilizes bidirectional and rich motion cues, allowing for continuous optimization for better results.}\label{fig:ranking}
\vspace{-10pt}
\end{figure}

However, the data captured by event cameras is spatially sparse, irregular, and asynchronous, making it unsuitable for directly applying current state-of-the-art image-based flow methods to estimate event-based optical flow. To leverage established convolutional neural network architectures~\cite{flownet, flownet2, raft}, recent works~\cite{evflownet,eraft} transform consecutive event streams into grid-like/tensor representations and input them into these networks to predict the optical flow. 

A representative method is E-RAFT~\cite{eraft}. Inspired by RAFT~\cite{raft}, E-RAFT employs image-based optical flow architectures for event-based optical flow estimation. It first converts two event sequences into grid-like/tensor representations, constructs a correlation volume between the two representations, and then regresses the optical flow from the correlation volume. The correlation volume, which measures visual similarity between pixels in two images, has become an essential component in mainstream image-based/frame-based optical flow estimation methods.  However, unlike dense and rich RGB or grayscale image data, event data is spatially sparse, resulting in the constructed correlation volume with insufficient motion cues, which significantly limits the performance of this approach. To provide more motion cues, 
TMA~\cite{tma} computes the correlation volumes between the reference event frame and multiple forward event frames.
Event data is inherently sequential. Although TMA has achieved improved results, it relies solely on forward temporal motion cues and overlooks backward temporal motion cues, which limits its overall performance (see Figure~\ref{fig:ranking}).


Our goal is to transform bidirectional temporally dense motion cues into spatially dense ones, thereby enabling detailed and accurate optical flow estimation. Our model is built on two key insights. First, events and motion are both temporally continuous and fine-grained, so forward and backward temporal motion cues are crucial for accurate optical flow estimation. Second, motion can vary non-uniformly over time, indicating the need to adaptively aggregate both forward and backward motion cues to ensure temporal consistency.

Building on these insights, we present bidirectional adaptive temporal (BAT) correlation for event-based optical flow estimation. Specifically, we split the reference event stream and target event stream into multiple groups, then compute bidirectional temporal correlation (BTC): forward correlation is applied between the reference event frame and multiple target event frames to obtain forward motion cues, while backward temporal correlation is performed between multiple reference event frames to capture backward motion cues. BTC transforms temporally dense motion cues into spatially dense ones, allowing for accurate and spatially dense optical flow estimation. 
We also propose an adaptive temporal sampling strategy for temporal consistency during computing correlation. 

Linearly indexing adjacent event frames across different time spans and performing correlation can introduce inconsistent motion patterns when dealing with non-uniformly moving objects. To address this challenge, we propose spatially adaptive temporal motion aggregation (SATMA). Specifically, SATMA first applies deformable and sparse attention between target motion features and adjacent motion features to aggregate relevant target motion features within the regions of interest. It then generates an attention map to adaptively fuse the aggregated target motion features into the adjacent motion features, effectively suppressing inconsistent motion features and enhancing consistent ones.

Extensive experiments demonstrate that our approach outperforms previous methods by a large margin on public benchmarks. Specifically, our BAT ranks $1^{st}$ on the DSEC-Flow~\cite{dsec} benchmark, exceeding E-RAFT~\cite{eraft} and TMA~\cite{tma} by 39.45\% and 28.98\%, respectively, for the 1PE metric. 

Additionally, the proposed backward temporal correlation offers two notable advantages: first, it enables us to predict future optical flow using only past events, a capability crucial for interpreting and interacting with dynamic environments, as shown in Figure~\ref{fig:backward_flow}; second, it effectively handles occlusions caused by objects moving out of the target frame, as shown in Figure~\ref{fig:occlusion}.



To summarize, our main contributions are as follows: 
\begin{itemize}
    \item We present a novel framework, BAT, that learns to estimate event-based optical flow with bidirectional adaptive temporal correlation. 
    \item We propose spatially adaptive temporal motion aggregation that efficiently integrates temporally consistent target motion features into adjacent motion features while suppressing inconsistent ones.
    \item Our BAT achieves state-of-the-art performance on the DSEC-Flow benchmark, outperforming all previously published methods.
    \item Our BAT can accurately predict future optical flow using only past events, while also effectively handling occlusions caused by objects moving out of the target frame.
\end{itemize}

\section{Related Work}
\label{sec:related}
Deep learning has demonstrated remarkable success in image matching tasks~\cite{igev,igev++,coatrsnet,monster,zerostereo,fast-acv,cheng2024adaptive}. The pioneering flow method, FlowNet~\cite{flownet}, introduced an end-to-end convolutional neural network that directly takes two images as input and outputs optical flow, achieving impressive performance. Subsequently, network architectures~\cite{raft,flowformer} and training strategies~\cite{sea-raft,sun2022disentangling} for optical flow estimation have been continuously improved, leading to consistent advancements in performance. For example, PWC-Net~\cite{pwcnet,sun2019models} and LiteFlowNet~\cite{liteflownet,lightweight2} introduced pyramid and warping-based cost volumes to estimate optical flow in a coarse-to-fine manner. These methods significantly reduce inference time while enhancing accuracy. RAFT presents a novel deep network architecture for optical flow, which first builds a 4D correlation volume for all pairs of pixels, and then iteratively regresses the optical flow from it. Combined with its new training strategy, RAFT sets a new state-of-the-art in optical flow estimation. 

Inspired by RAFT, several novel designs, such as global attention~\cite{gma}, kernel patch attention~\cite{kpaflow}, and super kernels~\cite{skflow}, have been proposed to improve the performance of motion aggregation. Recently, many works~\cite{flowformer,gmflow,transflow,craft,croco} have introduced transformers into optical flow estimation to enhance global matching. These methods have all led to improvements in accuracy. Efforts~\cite{flow1d,hcv,scv,meflow} have also been made to overcome the excessive memory consumption introduced by the high-resolution 4D correlation volume in RAFT. Instead of focusing on a two-frame setting, VideoFlow~\cite{videoflow} concurrently estimates bidirectional optical flows for multiple consecutive frames. In contrast to VideoFlow, our approach builds on the sparse nature of event data and transforms the rich temporal motion cues into the target frame, focusing on estimating its motion. UnFlow~\cite{unflow} designs an unsupervised loss on bidirectional optical flow computed from two static images. In contrast, our method computes bidirectional temporal correlation for more accurate and detailed forward optical flow prediction.


Learning-based event-based flow methods also largely draw upon well-established image-based optical flow architectures. Moreover, adapting to these architectures often requires converting events into tensor representations, like event frames or voxel grids. These learning-based methods can be categorized into supervised~\cite{eraft,tma,idnet}, semi-supervised~\cite{evflownet,spikeflownet,steflownet}, and unsupervised~\cite{tamingcm,paredes2021back,hagenaars2021self,zhu2019unsupervised}. Unsupervised methods rely solely on event data; semi-supervised methods use grayscale images as a supervisory signal; supervised methods require accurate ground truth flow for supervision. Currently, supervised methods dominate on standard benchmarks~\cite{dsec,mvsec}. A representative approach is E-RAFT~\cite{eraft}, which introduces the RAFT~\cite{raft} architecture into event-based optical flow estimation. Subsequently, TMA~\cite{tma} extends the two-frame correlation volume of E-RAFT to a multi-frame correlation volume, providing more temporal motion cues. ADMFlow~\cite{admflow} renders high-quality datasets to facilitate event-flow learning. Without using a correlation volume, IDNet~\cite{idnet} introduces iterative deblurring for accurate event-based flow estimation.

Contrary to existing learning-based event-based methods that only model unidirectional motion cues, we propose bidirectional adaptive temporal correlations to estimate spatially detailed optical flow. More importantly, by modeling backward motion cues, our method can predict future optical flow using only past events.

\section{Method}
\label{sec:method}

\begin{figure}
\centering
{\includegraphics[width=0.9\linewidth]{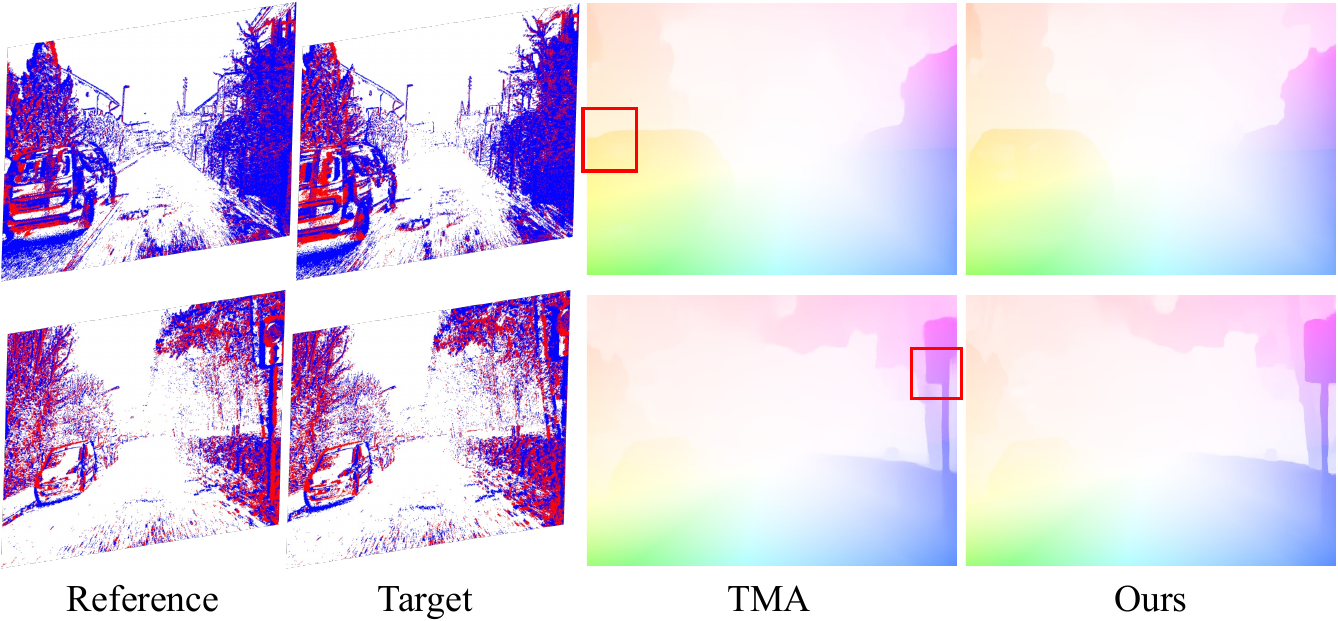}}
\caption{Benefiting from the proposed backward temporal correlation, our method effectively handles occlusions caused by objects moving out of the target frame.}\label{fig:occlusion}
\end{figure}

\begin{figure*}[t]
    \centering
    \includegraphics[width=0.9\linewidth]{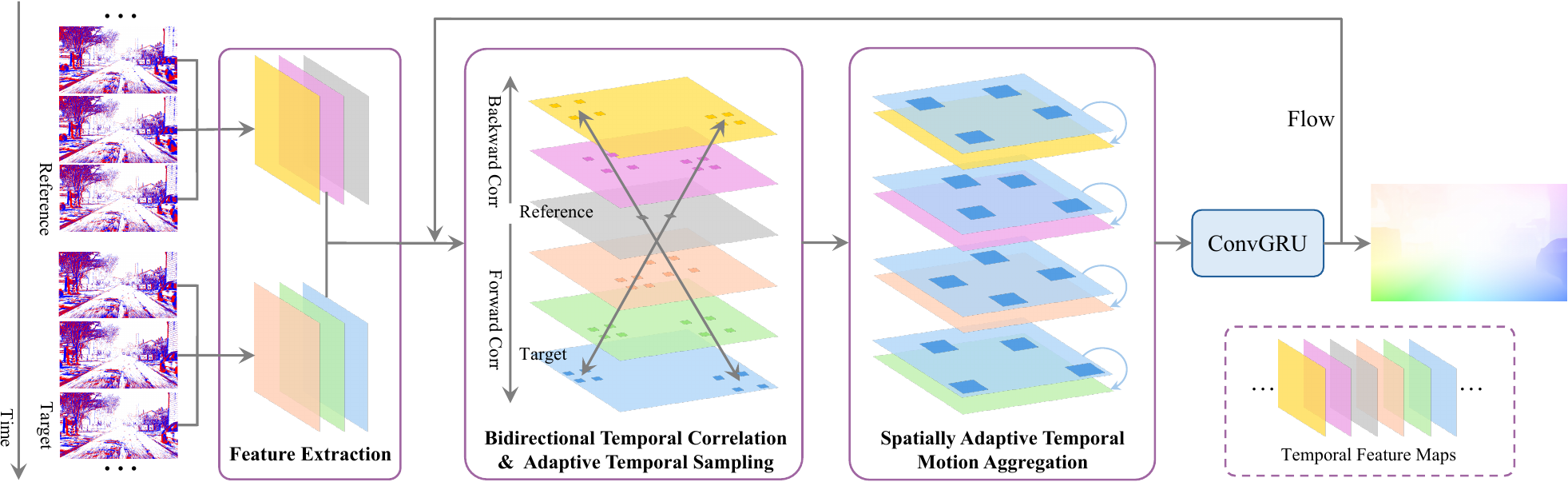}
    \caption{Network architecture of the proposed BAT. We first split the reference and target event streams into multiple groups and extract the corresponding features separately. Then, we perform forward and backward temporal correlations between the sequential event features. 
    We further propose spatially adaptive temporal motion aggregation, which integrates temporally consistent target motion features into adjacent motion features while suppressing inconsistent ones. 
    }
    \label{fig:network}
\end{figure*}

Conventional optical flow estimation is typically defined as finding pixel correspondences between two image frames. However, unlike frame-based images, event data is recorded in a sequential and asynchronous manner by detecting per-pixel brightness changes. Therefore, given an event stream $\mathcal{E}(t_i, t_{i+1})$, the goal of event-based optical flow is to find pixel correspondences between two timestamps $t_i$ and $t_{i+1}$.

Building on the spatial sparsity and temporal density of event data, we propose BAT, a novel framework that transforms temporally dense motion cues into spatially dense ones, enabling spatially dense and accurate optical flow estimation. The overview of the proposed network framework is illustrated in Figure~\ref{fig:network}. 
Before feeding the event data into the network, the event stream is first converted into an event voxel grid representation. In this section, we first describe the event voxel grid representation and then detail the network architecture.

\subsection{Event Representation}
Event cameras are novel bio-inspired vision sensors that generate an asynchronous event stream when the brightness change of a pixel exceeds a predefined threshold $C$. An event $e_k(t) = (x_k, y_k, t, p_k)$ is a tuple containing trigger pixel location ($x_k$, $y_k$), the timestamp $t_k$ and the polarity $p_k$ that indicates whether the brightness change is positive or negative. The event stream $\mathcal{E}(t_{i}, t_{i+1})$ from timestamp $t_i$ to timestamp $t_{i+1}$ is defined as ${\mathcal{E}}(t_{i}, t_{i+1}) := \langle e(t) | t \in [t_i, t_{i+1}]\rangle$. 

To ensure compatibility with CNNs and facilitate correlation volume construction, we follow previous works~\cite{eraft,tma} by transforming the event stream $\mathcal{E}(t_{i}, t_{i+1})$ into an event voxel grid $\bm {V} \in \mathbb{R}^{B \times H_0 \times W_0}$ with discretized time dimension $B$. Specifically, given an event stream $\mathcal{E}(t_{i}, t_{i+1})$, $\bm {V}(b, x, y)$ is obtained by:
\begin{equation}
\begin{aligned}
&t_{j}^{*} = (B-1)(t_{j}-t_{i}) /(t_{i+1}-t_{i}) \\
&\bm {V}(b, x, y) = \sum_{j} p_{j} k_{b}(x-x_{j}) k_{b}(y-y_{j}) k_{b}(b-t_{j}^{*}) \\
&k_{b}(a) = \max (0,1-|a|),
\end{aligned}
\label{equ:evevt_voxel}
\end{equation}

where $B$ denotes the number of time bins, and $b \in [0, B-1]$ is the integer bin index.

\subsection{Feature Extraction}
\label{sec:fea_extra}
Given two consecutive event streams $\mathcal{E}(t_{i-1}, t_{i})$ and $\mathcal{E}(t_{i}, t_{i+1})$, where $t_{i+1} - t_i = t_{i} - t_{i-1} = \Delta t$, our goal is to estimate the optical flow $\bm f_{t_i\rightarrow{t_{i+1}}}$ form timestamp $t_i$ to $t_{i+1}$. Firstly, we convert the event streams into voxel grid representations, denoted as $\bm {V}_0$ and $\bm {V}_1$, respectively, using Equ.~\ref{equ:evevt_voxel}. 

To exploit temporally dense motion cues, we split both $\bm {V}_0$ and $\bm {V}_1$ into $N$ groups equally along the time bin dimension, and each group contains $B/N$ time bins. Then we input the voxel grid event representations of these 2N groups into a shared-weight feature extraction network to produce a series of features $\bm F_n$ ($\bm F_n \in \mathbb{R}^{D \times H \times W}$), $n=1,2,\dots, 2N$, where $D$, $H$ and $W$ denote the channel, height, and width of feature $\bm F_n$. The feature extraction network consists of 6 residual blocks, similar to RAFT~\cite{raft}. Among these temporal features, $\bm F_{N}$ serves as the reference frame, while $\bm F_{2N}$ serves as the target frame.


\subsection{Bidirectional Temporal Correlation}
Correlation volume, which measures pixel-wise visual similarities between frames, provides crucial information for flow estimation. However, existing flow methods often focus on the two-frame setting~\cite{eraft} or unidirectional correlation~\cite{tma}. In contrast, we propose bidirectional temporal correlation (BTC) to transform temporally rich motion cues into spatially dense ones. BTC consists of forward temporal correlation and backward temporal correlation. Forward temporal correlation computes the correlation between the reference frame $\bm F_{N}$ and its next frames $\bm F_{N+j}$ ($j=1,\dots, N$), and backward temporal correlation computes the correlation between the reference frame $\bm F_{N}$ and its previous frames $\bm F_{N-j}$ ($j=1,2,\dots, N-1$).

Given the current flow estimate $\bm f$ from timestamp $t_i$ to $t_{i+1}$, in order to facilitate the computation of temporal correlations, we assume that the optical flow is linear with respect to time. Thus, we can derive the optical flow $\bm {df}$ between adjacent temporal groups as:
\begin{equation}
\begin{aligned}
    {\bm {df}} = {\bm f}/N
\end{aligned}
\end{equation}
Then, the forward temporal correlation ${\bm C}_j^{fwd} \in \mathbb{R}^{(2r+1)^2 \times H \times W}$ is computed by:
\begin{equation}
\begin{aligned}
{\bm C}_j^{fwd} = \frac{{\bm F}_{N} \boldsymbol{\cdot} \mathcal{W}({\bm F}_{N+j};  j \bm {df}) }{\sqrt{D}}, j=1,2,\dots,N,
\end{aligned}
\end{equation}
where $\boldsymbol{\cdot}$ denotes the dot product, $\mathcal{W}(\cdot; \cdot)$ denotes the warping of adjacent frames to the reference frame using optical flow. Specifically, for each pixel $p$ in 
$\bm F_N$, we use the current flow $j \bm {df}$ to find its corresponding point $p'$ in $\bm F_{N+j}$, then define a local grid around $p'$ within a radius 
$r$ and bilinearly sample the values from the local grid. The number of samples for the local grid is $(2r+1)^2$.

Similarly, the backward temporal correlation $\bm {C}_j^{bwd} \in \mathbb{R}^{(2r+1)^2 \times H \times W}$ is computed by:
\begin{equation}
\begin{aligned}
\bm {C}_j^{bwd} = \frac{\bm {F}_{N} \boldsymbol{\cdot} \mathcal{W}(\bm {F}_{N-j}; (-j)\bm {df}) }{\sqrt{D}}, j=1,2,\dots,N-1.
\end{aligned}
\end{equation}

Finally, we obtain $N$ groups of forward temporal correlation maps and $N\!-\!1$ groups of backward temporal correlation maps. Through this bidirectional temporal correlation, we transform temporally dense motion cues into spatially dense ones, enabling spatially detailed and accurate optical flow estimation. In particular, the backward temporal correlation maps are highly beneficial for handling occlusions of objects moving out of the target frame, as shown in Figure~\ref{fig:occlusion}.

\textbf{Adaptive Temporal Sampling Strategy}. The sampling radius 
$r$ is set as a hyperparameter in previous works~\cite{eraft,tma}. However, as motion patterns change over time, a manually chosen $r$ may not be robust for temporal consistent correlation. To address this, we introduce a learnable parameter $\alpha$ as a scale factor to obtain a learnable sampling radius $lr$,
\begin{equation}
\begin{aligned}
lr = \alpha \cdot r.
\end{aligned}
\end{equation}
The $lr$  can be adaptively learned during the training process, while the number of sampling points in the local grid remains unchanged.

\subsection{Spatially Adaptive Temporal Motion Aggregation}
\label{sec:satma}
Effectively utilizing temporally rich motion cues is crucial for achieving spatially dense and accurate optical flow estimation. However, due to the non-uniformity of optical flow over time, linearly computing correlations on adjacent temporal frames introduces motion features inconsistent with the target frame. 
To address this, we propose a spatially adaptive temporal motion aggregation (SATMA) module that adaptively enhances motion features consistent with the target frame while suppressing inconsistent ones. The architecture of SATMA is illustrated in the supplementary material.

Before feeding into STAMA, we encode the correlation feature $\bm {C}_j^{fwd}$ into motion feature $\bm {M}_j^{fwd}$:
\begin{equation}
\begin{aligned}
\bm {M}_j^{fwd} = \text{MotionEncoder}({\bm C}_j^{fwd}, \bm f), j=1,2,\dots,N,
\end{aligned}
\end{equation}
Similarly, we can obtain the backward motion feature $\bm {M}_j^{bwd}$ ($j=1,2,\dots,N-1$). Among them, $\bm {M}_N^{fwd}$ represents the target motion feature, while the others are adjacent motion features.

The architecture of the spatially adaptive temporal motion aggregation (SATMA) module is presented in the supplementary material. Given the target motion feature $\bm {M}_N^{fwd}$ and an adjacent motion feature $\bm {M}_j^{fwd}$ (or $\bm {M}_j^{bwd}$), we first concatenate them, then pass them through a convolutional layer followed by a sigmoid activation function to generate a spatial attention map $ \bm A_{spa}$. The spatial attention effectively enhances consistent motion features while suppressing inconsistent ones. We also adopt deformable attention~\cite{deform_attn} to efficiently integrate consistent or similar motion features from $\bm {M}_N^{fwd}$ into $ \bm {M}_j^{fwd}$. Specifically, $\bm {M}_j^{fwd}$ is first projected linearly to obtain the query $\bm Q$, which is then fed into a lightweight offset network to generate sparse sampling locations. Based on these locations, we can sample $\bm {M}_N^{fwd}$ to obtain the key $\bm K$ and value $\bm V$. Then we obtain the aggregated motion feature $\bm {M}^{agg}_j$ through attention:
\begin{equation}
\begin{aligned}
\bm {M}^{agg}_j = \sigma(\frac{\bm Q \bm K^T}{\sqrt{D_m}})\bm V,
\end{aligned}
\end{equation}
where $\sigma(\cdot)$ denotes the softmax function, and $D_m$ is the dimension of the motion feature. Finally, we obtain fused motion feature $\bm {M}^{fuse}_j$ by:
\begin{equation}
\begin{aligned}
\bm {M}^{fuse}_j = \bm A_{spa} \odot \bm {M}^{agg}_j + \bm {M}_j^{fwd},
\end{aligned}
\end{equation}
where $\odot$ denotes the element-wise product. Similarly, we also perform SATMA between backward motion features and the target motion feature to obtain the backward fused motion features. Then, these forward and backward fused motion features are fed into the ConvGRU to update the optical flow.

\begin{table}
\setlength{\tabcolsep}{3pt}
\centering 
\begin{tabular}{llccccc}
\toprule
 &Method & EPE$\downarrow$ & 1PE$\downarrow$ & 2PE$\downarrow$ & 3PE$\downarrow$ & AE$\downarrow$ \\
\midrule
MB & MultiCM &3.472 & 76.570 & 48.480 & 30.855 & 13.983 \\
\midrule
\multirow{6}{*}{SL} & E-RAFT & 0.788 & 12.742 & 4.740 & 2.684 & 2.851 \\
& ADMFlow & 0.779 & 12.522 & 4.673 & 2.647 & 2.838 \\
& EEMFlow+ & 0.751 & 11.403 & 3.932 & 2.145 & 2.669 \\
& TMA &0.743 & 10.863 & 3.972 & 2.301 & 2.684 \\
& IDNet & 0.719 & 10.069 & 3.497 & 2.036 & 2.723 \\
& BAT (Ours) & \textbf{0.655} & \textbf{7.715} & \textbf{2.896} & \textbf{1.773} & \textbf{2.439} \\
\bottomrule
\end{tabular}
\caption{DSEC~\cite{dsec} benchmark evaluation. MB refers to model-based methods that do not require training data, and SL represents supervised learning methods that require ground truth optical flow. \textbf{Bold}: best.}
\label{tab:dsec_benchmark}
\end{table}


\subsection{Loss Function}
Following RAFT, we supervised our network using the $l_1$ loss between the predicted and ground truth flow over the entire sequence of predictions, \{$\bm f^1, \bm f^2, \cdots, \bm f^{K}$\}, applying exponentially increasing weights. Given the ground truth flow $\bm f^{gt}$, the loss is defined as,
\begin{equation}
    \mathcal{\bm L} = \sum_{i=1}^{K} \gamma^{K-i} ||{\bm f}^i-{\bm f}^{gt}||_1,
\end{equation}
where $\bm f^i$ denotes the predicted optical flow at the $i$-th iteration, and $\gamma$ balances the weights of the loss terms for each iteration. $K$ is the total number of iterations.

\section{Experiments}
\label{sec:experiment}

\begin{table}
  \centering
    \setlength{\tabcolsep}{3.5pt}
  \begin{tabular}{llcccc}
    \toprule
     &\multirow{2}{*}{Method}  & \multicolumn{2}{c}{$dt = 1$} & \multicolumn{2}{c}{$dt = 4$} \\ 
     \cmidrule(lr){3-4} \cmidrule(lr){5-6}
     & & EPE$\downarrow$ & \%Out$\downarrow$ & EPE$\downarrow$ & \%Out$\downarrow$ \\
\midrule
MB & MultiCM & 0.30 & 0.10 & 1.25 & 9.21 \\
\midrule
\multirow{3}{*}{SSL} & EV-FlowNet & 0.49 & 0.20 & 1.23 & 7.30 \\
& Spike-FlowNet & 0.49 & - & 1.09 & - \\
& STE-FlowNet & 0.42 & 0.00 & 0.99 & 3.90 \\
\midrule
\multirow{6}{*}{SL} & E-RAFT & 0.27 & 0.00 & 0.84 & 1.70 \\
& DCEIFlow & 0.91 & 0.71 & 1.87 & 19.1\\
& ADMFlow & 0.52 & 0.00  & 1.91 & 19.2 \\
& TMA & 0.25  & 0.07 & 0.70 & 1.08 \\
& IDNet & 0.29 & 0.00 & 0.75 & 1.20 \\
& BAT (Ours) & \textbf{0.21} & 0.00 & \textbf{0.53} & \textbf{0.71} \\
\bottomrule
  \end{tabular}
  \caption{
  MVSEC~\cite{mvsec} benchmark evaluation. SSL denotes semi-supervised learning methods that use grayscale images for supervision. }
  \label{tab:mvsec_benchmark}
  \vspace{-10pt}
\end{table}


\begin{table}
  \centering
  \begin{tabular}{l|ccccc}
    \toprule
    Model & EPE$\downarrow$ & 1PE$\downarrow$ & 2PE$\downarrow$ & 3PE$\downarrow$ & AE$\downarrow$ \\ 
\midrule
Baseline & 0.712 & 9.123 & 3.332 & 2.033 & 2.544 \\
\midrule
BTC & 0.680 & 8.279 & 3.130 & 1.889 & 2.527 \\
BTC+ATS  & 0.671 & 8.179 & 3.014 & 1.804 & 2.482 \\
Full (BAT) & \textbf{0.655} & \textbf{7.715} & \textbf{2.896} & \textbf{1.773} & \textbf{2.439} \\
    \bottomrule
  \end{tabular}
  \caption{
  Ablation study on DSEC-Flow benchmark. BTC represents bidirectional temporal correlation, ATS represents adaptive temporal sampling. Full model includes SATMA, denoting spatially adaptive temporal motion aggregation.}
  \label{tab:ablation}
\end{table}

\begin{table}
  \centering
  \setlength{\tabcolsep}{5pt}
  \begin{tabular}{llccc}
    \toprule
    Experiment & Method & EPE$\downarrow$ & 1PE$\downarrow$ & 3PE$\downarrow$
    \\
     \midrule
    \multirow{3}{*}{Attention} & dense & 0.670 & 8.049 & 1.868 \\
     & spatial-reduction & 0.698 & 8.731 & 2.044 \\
     & \underline{deformable} & \textbf{0.655} & \textbf{7.715} & \textbf{1.773} \\
     \midrule
     \multirow{3}{*}{Radius} & r=0 & 0.677 & 8.02 & 1.985 \\
     & r=1 & 0.667  & 7.981 & 1.945 \\
     & \underline{r=2} & \textbf{0.655} & \textbf{7.715} & \textbf{1.773} \\
     \midrule
     \multirow{3}{*}{Iterations} & 4 & 0.689 & 8.563 & 2.011 \\
     & \underline{8} & 0.655 & 7.715 & 1.773 \\
     & 12 & \textbf{0.650} & \textbf{7.540} & \textbf{1.740} \\    
    \bottomrule
  \end{tabular}
  \caption{
  Ablation experiments on DSEC-Flow benchmark. Settings used in our final model are \underline{underlined}.}
  \label{tab:ablation_v2}
\end{table}


\begin{table}
\centering 
\setlength{\tabcolsep}{4pt}
\begin{tabular}{llcc}
\toprule
 Input & Method & EPE$\downarrow$ & 1PE$\downarrow$ \\
\midrule
\multirow{3}{*}{Tgt.+Ref. event}
 & MultiCM &3.472 & 76.570 \\
 & TamingCM & 2.330 & 68.293 \\
 & EV-FlowNet & 2.320 & 55.400 \\
\midrule
\multirow{2}{*}{only Ref. event}
& E-RAFT (warm-start) & 4.518 & 85.378 \\
& BAT (bwd corr) & \textbf{1.163} & \textbf{33.026} \\
\bottomrule
\end{tabular}
\caption{Future optical flow prediction. Ref. denotes reference event stream $\mathcal{E}(t_{i-1}, t_{i})$, and Tgt. denotes target event stream $\mathcal{E}(t_{i}, t_{i+1})$. Our BAT (bwd corr) can predict future optical flow $\bm f_{t_i\rightarrow{t_{i+1}}}$ using only the past event 
 stream $\mathcal{E}(t_{i-1}, t_{i})$.}
\label{tab:future_flow}
\end{table}

\subsection{Experimental Setup}

\noindent\textbf{Datasets and evaluation setups.} Following previous work, we conduct extensive experiments on two popular event-based datasets: DSEC-Flow~\cite{dsec} and MVSEC~\cite{mvsec}. For DSEC-Flow, we train the models on the official training dataset and then evaluate them on the public benchmark. DSEC-Flow provides 8,170 training samples and 416 testing samples, with a resolution of $640\times480$. For MVSEC, we perform training and evaluation on two types of event data, $dt=4$ and $dt=1$, which are divided based on time intervals.


\vspace{2mm}
\noindent\textbf{Implementation details.} 
We implement our model with PyTorch and perform our experiments using NVIDIA RTX 3090 GPUs. We set the temporal group $N$ to 3 in the final BAT model. For experiments on DSEC-Flow, we set the event representation to 
$B=15$ time bins for every 100ms of events. For MVSEC, 
$B$ is set to 5 when $dt=1$ and 15 when $dt=4$. The length $B$ of time bins is kept consistent with E-RAFT and TMA. The final channel dimension 
$D$ for feature extraction is 128, and the flow updater is identical to that of RAFT. The sampling radius $r$ is set to 2. For all training, we use the AdamW optimizer with a one-cycle learning rate schedule, setting the learning rate to 2e-4. We use 8 update iterations during training, and the weight $\gamma$ is set to 0.8. For DSEC-Flow, we train our BAT for 200k steps with a batch size of 8, while for MVSEC, we train it for 100k steps.

\begin{figure}
\centering
{\includegraphics[width=0.9\linewidth]{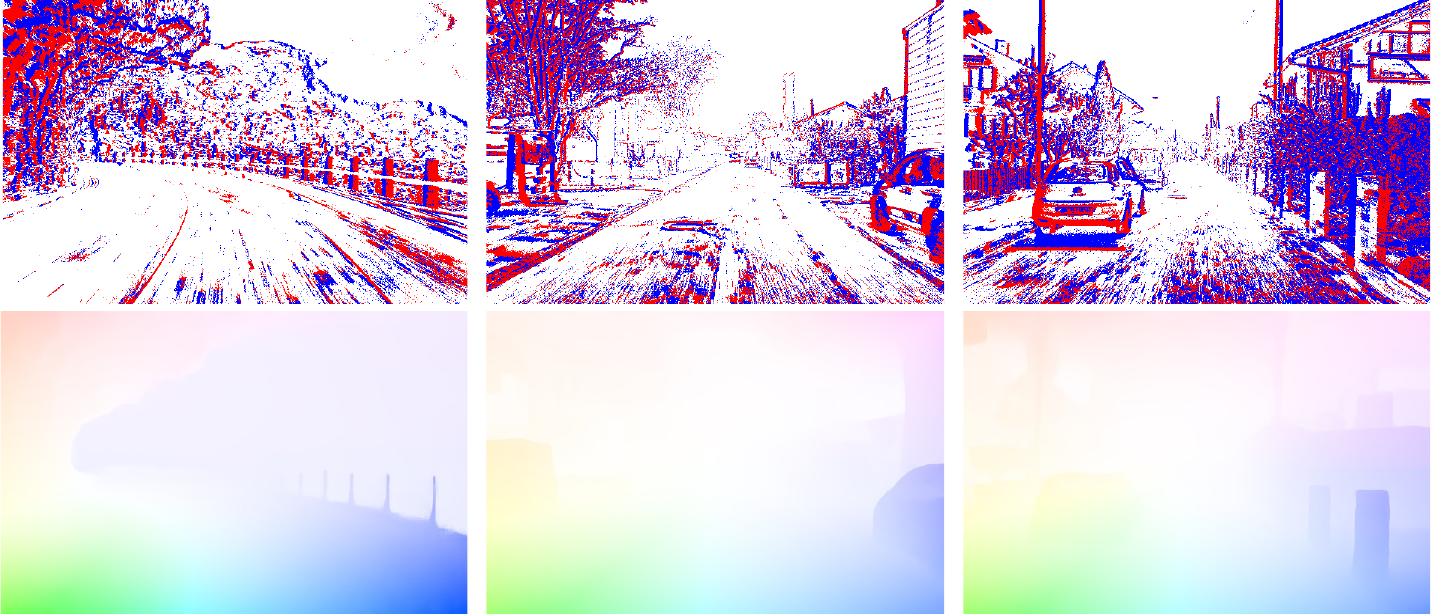}}
\caption{Future optical flow prediction. The first row illustrates events from timestamp $t_{i-1}$ to $t_i$, while the second row presents the optical flow results from $t_{i}$ to $t_{i+1}$. Given only past events, our method can predict future optical flow.}\label{fig:future_flow}
\end{figure}

\subsection{Comparisons with State-of-the-art}
\textbf{DSEC-Flow.} Table~\ref{tab:dsec_benchmark} presents the comparison results on the DSEC-Flow benchmark. Our BAT ranks $1^{st}$ among all published methods for all metrics. 
Compared to other iterative methods, E-RAFT~\cite{eraft} and TMA~\cite{tma}, our BAT improves the 1PE metric by 39.45\% and 28.98\%, respectively. The 1PE metric is the most challenging evaluation metric, as it directly reflects the detailed performance of the predictions. 

\begin{figure*}
\centering
\includegraphics[width=0.75\textwidth]{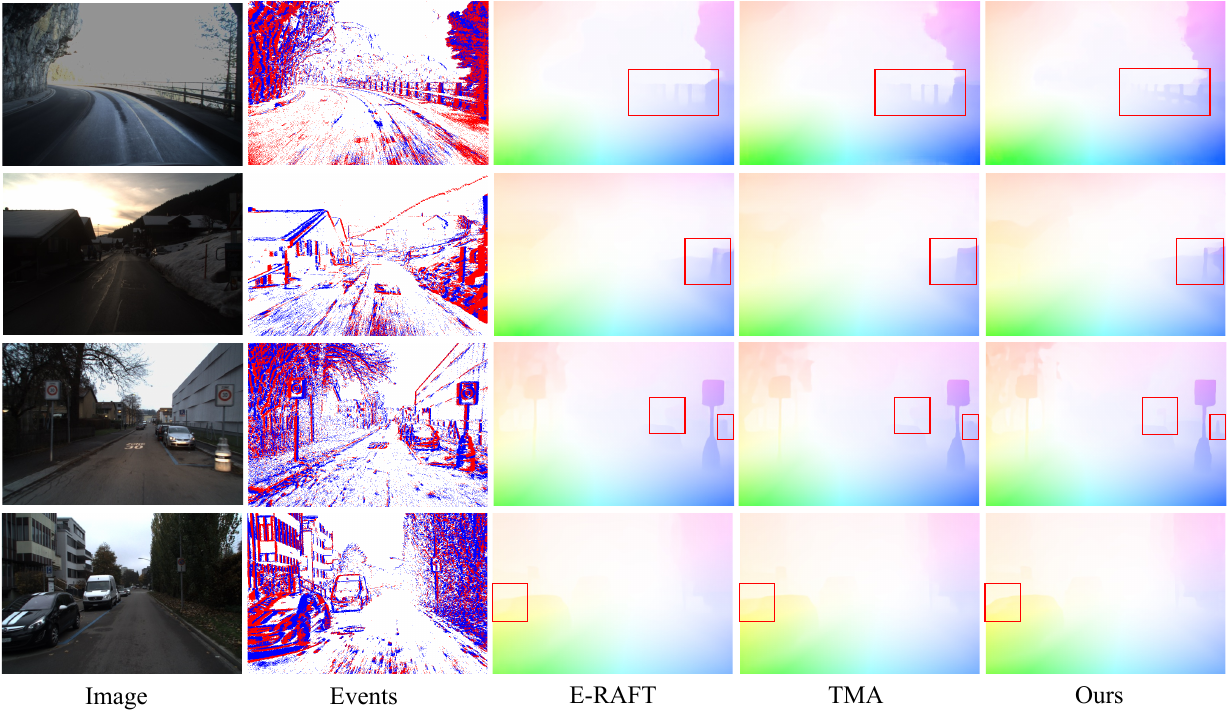} 
\caption{Qualitative results of optical flow predictions on DSEC-Flow. Significant improvements are highlighted by red boxes. Images are provided for visualization only, since the optical flow is event-based. Our method accurately distinguishes subtle details and sharp edges. 
Zoom in for a clearer view.}
\vspace{-10pt}
\label{fig:dsec}
\end{figure*}

Qualitative results on the DSEC-Flow benchmark are shown in Figure~\ref{fig:dsec}. Due to the spatial sparsity of event data, previous methods~\cite{eraft,tma} tend to produce predictions with blurred edges and loss of details. In contrast, our BAT leverages bidirectional temporal motion cues, enabling it to predict sharp edges and preserve fine structures. Additionally, benefiting from the proposed backward temporal correlation, our BAT effectively handles occlusions caused by objects moving out of the target frame, as shown in Figure~\ref{fig:occlusion}.

\vspace{2mm}
\noindent\textbf{MVSEC.} Following previous works, we perform training on outdoor\_day2 sequence and evaluate on outdoor\_day1 sequence. Table~\ref{tab:mvsec_benchmark} presents the evaluation results corresponding to $dt=1$ grayscale frame and $dt=4$ grayscale frame. Our method achieves the highest prediction accuracy and outperforms other methods by a large margin. Specifically, for the more challenging $dt=4$ setting, our BAT surpasses the second-best method, TMA~\cite{tma}, by 34.26\% for the \%Out metric and 24.29\% for the EPE metric. 

\begin{figure}
\centering
{\includegraphics[width=1.0\linewidth]{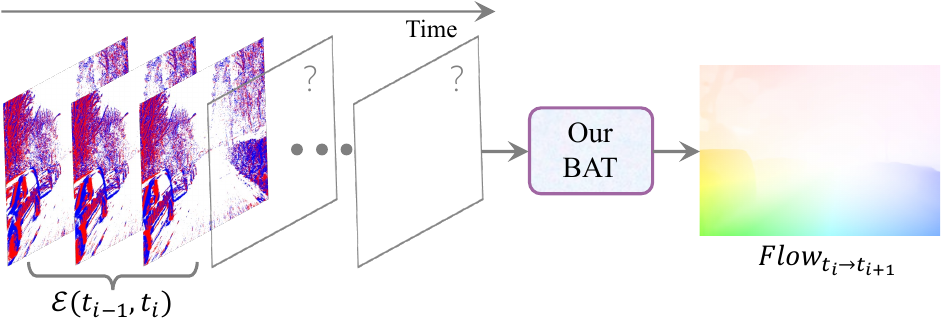}}
\caption{Predicting future optical flow $\bm f_{t_i\rightarrow{t_{i+1}}}$ using only past events $\mathcal{E}(t_{i-1}, t_{i})$.}\label{fig:backward_flow}
\vspace{-10pt}
\end{figure}

\subsection{Ablation Study}
We conduct ablation studies to validate the effectiveness of the proposed components. All ablation models are trained on the DSEC-Flow training dataset and evaluated on the DSEC-Flow benchmark, as presented in Table~\ref{tab:ablation}. We use the TMA~\cite{tma} with $N=3$ temporal groups as the Baseline. Baseline only computes forward temporal correlation, neglecting the backward motion cues. In contrast, the proposed bidirectional temporal correlation (BTC) computes both forward and backward temporal correlations, enabling the model to leverage bidirectional temporally dense motion cues for accurate and spatially detailed optical flow estimation. As a result, BTC achieves a significant improvement in accuracy. Furthermore, we introduce an adaptive temporal sampling (ATS) strategy to ensure temporal consistency when computing correlations, which enhances the model's performance.

Due to the presence of non-uniform motion, simply sampling the non-target event frames based on a linear motion assumption can introduce inconsistent temporal motion features. To mitigate this issue, we propose spatially adaptive temporal motion aggregation (SATMA), which efficiently and adaptively aggregates consistent target motion features while suppressing inconsistent motion features. As shown in Table~\ref{tab:ablation}, SATMA significantly improves the model's prediction accuracy.


\noindent\textbf{Attention type.} To perform temporal motion aggregation effectively, we compare three attention types (Table~\ref{tab:ablation_v2}). Dense attention incurs high memory/computation and introduces irrelevant features. Spatial-reduction attention lowers the cost by pooling $\bm K$ and $\bm V$, but sacrifices detail and degrades accuracy. In contrast, deformable attention efficiently focuses on relevant motion features with low overhead.

\noindent\textbf{Sampling radius.} Due to the rich temporal motion cues, even with using a sampling radius of 0, computing correlation at a single point, our method can still yield good results, as shown in Table~\ref{tab:ablation_v2}. Better results are obtained with an increased radius.

\noindent\textbf{Iterations.} By effectively integrating bidirectional and rich temporal motion cues, our results continuously improve with more iterations, as shown in Table~\ref{tab:ablation_v2} and Figure~\ref{fig:ranking}.

\subsection{Future Optical Flow Prediction}
Predicting future optical flow is a crucial task for interpreting and interacting with dynamic environments, such as assisting drones or autonomous vehicles in navigating through complex and dynamic environments. Previous methods~\cite{evflownet,eraft,tma} require the given $\mathcal{E}(t_{i}, t_{i+1})$ to predict flow $\bm f_{t_i\rightarrow{t_{i+1}}}$, making them difficult to predict future optical flow. Benefiting from the proposed backward temporal correlation, our model can predict future optical flow $\bm f_{t_i\rightarrow{t_{i+1}}}$ using only the past event stream $\mathcal{E}(t_{i-1}, t_{i})$, shown in Figure~\ref{fig:backward_flow}.
Our BAT (bwd corr) produces promising optical flow results on the DSEC-Flow benchmark, significantly outperforming the model-based method MultiCM, the self-supervised method TamingCM, and E-RAFT’s warm-start approach that initializes the flow for the next frame, as shown in Table~\ref{tab:future_flow}. Visual results are shown in Figure~\ref{fig:future_flow}.

\section{Conclusion}

In this paper, we present BAT, an innovative framework designed to estimate accurate and detailed optical flow from events. BAT introduces three key innovations: a bidirectional temporal correlation, an adaptive temporal sampling strategy, and spatially adaptive temporal motion aggregation. Our approach ranks $1^{st}$ on the DSEC-Flow benchmark, outperforming existing state-of-the-art methods by a large margin, while also preserving sharp edges and high-quality details. Additionally, our BAT can accurately predict future optical flow using only past events. We believe that our simple method, BAT, will become the baseline for the community and drive the future development of event-based optical flow.

\noindent\textbf{Limitations.} Our BAT does not bring significant gains under rapid motion changes. For example, when the event camera shakes violently, there is a significant difference between backward and forward temporal motion. In this case, backward temporal motion cues provide little help for forward optical flow estimation.

\section{Acknowledgments}
This research is supported by the National Key R\&D Program of China (2024YFE0217700), the National Natural Science Foundation of China (623B2036,62472184), the Fundamental Research Funds for the Central Universities, and the Innovation Project of Optics Valley Laboratory (Grant No. OVL2025YZ005).

\bibliography{aaai2026}

\newpage

\begin{figure}
\centering
{\includegraphics[width=1.0\linewidth]{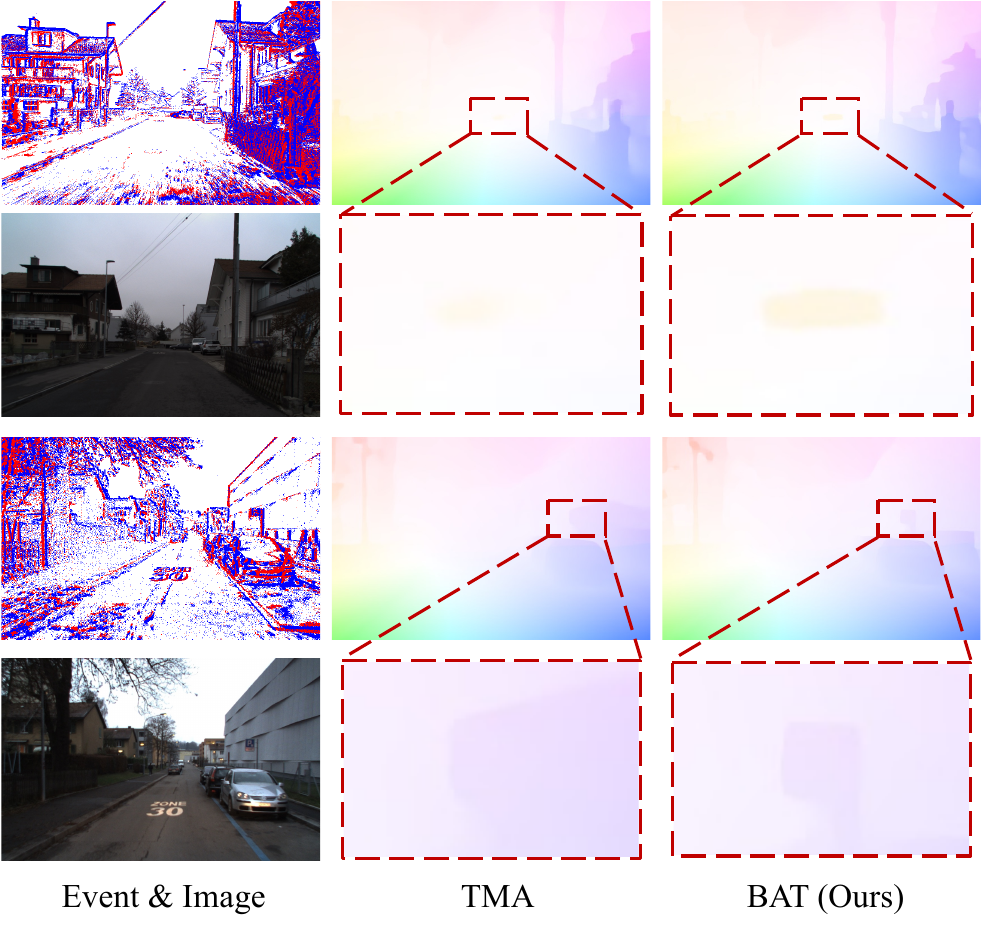}}
\caption{More qualitative results on DSEC-Flow~\cite{dsec}. Significant improvements are highlighted by red boxes. Images are provided for visualization only, since the optical flow is event-based. Our BAT effectively captures subtle details, while TMA~\cite{tma} struggles to predict them accurately.}
\label{fig:more_visual}
\end{figure}

\begin{figure}
\centering
{\includegraphics[width=1\linewidth]{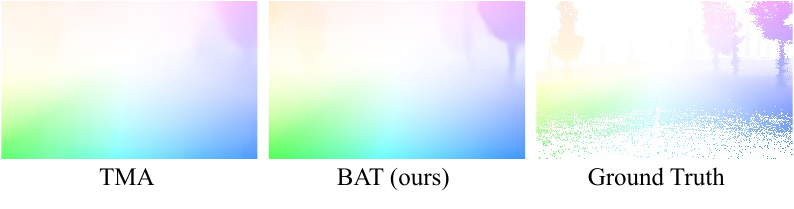}}
\caption{Qualitative comparisons with TMA on MVSEC.}\label{fig:mvsec}
\end{figure}

\begin{figure*}
    \centering
    \includegraphics[width=0.75\linewidth]{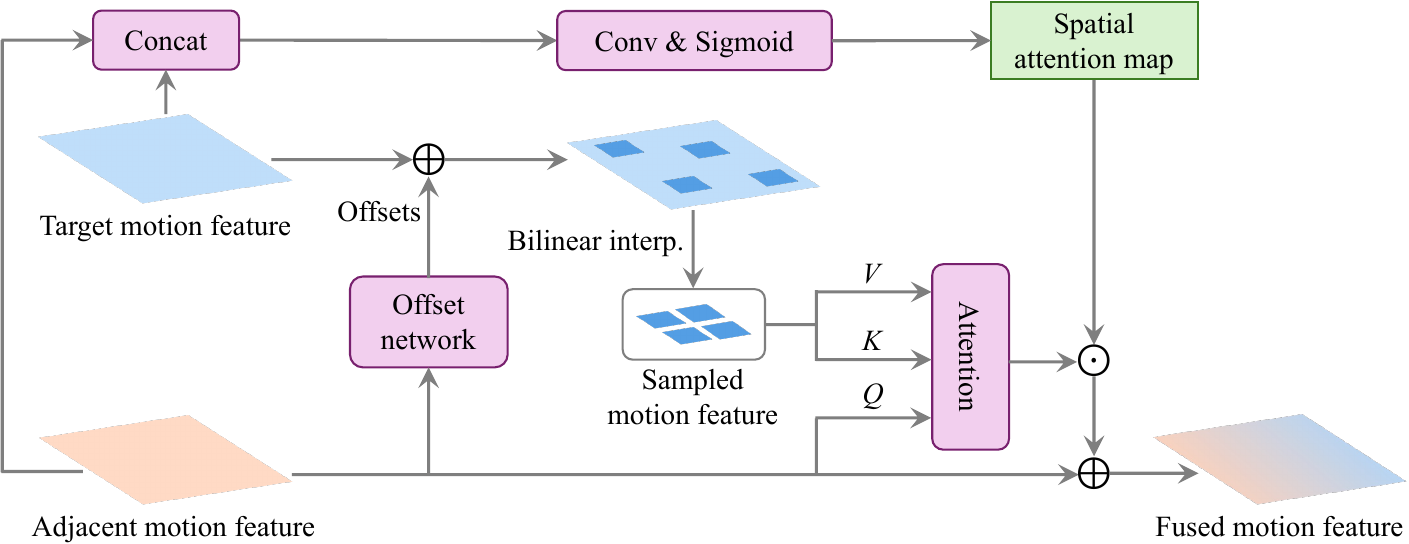}
    \caption{Spatially Adaptive Temporal Motion Aggregation (SATMA). 
    }
    \label{fig:satma}
\end{figure*}

\section{Details of Iterative Update}
After performing the spatially adaptive temporal motion aggregation, we obtain forward fused motion features and backward fused motion features. These motion features are then concatenated to form the overall bidirectional motion features, represented as:
\begin{equation}
\begin{aligned}
\bm {M}^{bid} =  \text{Concat}(\bm {M}^{fwd}, \bm {M}^{bwd}),
\end{aligned}
\end{equation}
where $\bm {M}^{fwd}$ and $\bm {M}^{bwd}$ denote the forward and backward fused motion features, respectively, $\bm {M}^{bid}$ denotes bidirectional overall motion features. $\text{Concat}(\cdot,\cdot)$ represents the concatenation operation.

In addition to the bidirectional temporal motion features $\bm {M}^{bid}$, we also use a context network to extract context features $\bm {C}_{F}$. The context network takes event frames from timestamp $t_i-\Delta t/N$ to $t_{i+1}$ as input, and its architecture is identical to that of the feature extraction network. We then concatenate the $\bm {M}^{bid}$ and $\bm {C}_{F}$ to form $x_k$. Finally, the iterative update is performed using two ConvGRUs: one with a $1\times5$ convolution and one with a $5\times1$ convolution,
\begin{equation}
\begin{aligned}
z_k = & \;\sigma(\text{Conv}_{1\times5}([h_{k-1}, x_k], W_z)), \\
r_k = & \;\sigma(\text{Conv}_{1\times5}([h_{k-1}, x_k], W_r)), \\
\Tilde{h}_k = & \,\tanh(\text{Conv}_{1\times5}([r_k \odot h_{k-1}, x_k], W_h)), \\
h_k = & \;(1-z_k) \odot h_{k-1} + z_k \odot \Tilde{h}_k, \\
z_k = & \;\sigma(\text{Conv}_{5\times1}([h_k, x_k], W_z)), \\
r_k = & \;\sigma(\text{Conv}_{5\times1}([h_k, x_k], W_r)), \\
\Tilde{h}_k = & \,\tanh(\text{Conv}_{5\times1}([r_k \odot h_k, x_k], W_h)), \\
h_k = & \;(1-z_k) \odot h_k + z_k \odot \Tilde{h}_k,
\end{aligned}
\end{equation}
where $\sigma$ denotes the $sigmoid$ function, $W_z$, $W_r$, and $W_h$ are the parameters of the network. Based on the hidden state $h_k$, we decode a residual optical flow $\Delta \bm f$ through two convolutional layers. The $\Delta \bm f$ is added to the current optical flow estimation to update the optical flow.

\section{More Visual Results}
\subsection{More comparisons with previous methods}
As shown in Figure~\ref{fig:more_visual}, we present additional qualitative results to further demonstrate the superiority of BAT. Compared with TMA~\cite{tma}, our BAT captures subtle details more effectively.

\subsection{Qualitative comparisons on MVSEC}
We provide visual comparisons with TMA~\cite{tma} on the MVSEC~\cite{mvsec} dataset. Our method produces high-quality flow results while preserving fine structures, as shown in Figure~\ref{fig:mvsec}.

\end{document}